\pdfoutput=1

\documentclass[11pt]{article}

\usepackage[final]{acl}

\usepackage{times}
\usepackage{booktabs}

\usepackage{amsmath}
\usepackage{amsfonts}
\usepackage{paralist}
\usepackage{latexsym}
\usepackage{subfig}
\usepackage{soul}
\usepackage{comment}
\usepackage{multicol}
\usepackage{multirow}
\usepackage{hhline}
\usepackage{float}
\usepackage[T1]{fontenc}

\usepackage[utf8]{inputenc}

\usepackage{microtype}

\usepackage{inconsolata}

\usepackage{graphicx}
\usepackage{subcaption}

\usepackage{todonotes} 

\usepackage{tabularray}
\newcommand{\sys}{\emph{MAPLE}}

\title{MAPLE: A Meta-learning Framework for Cross-Prompt Essay Scoring
}

\author{
  \textbf{Salam Albatarni}, \textbf{May Bashendy}, \textbf{Sohaila Eltanbouly}, \textbf{Tamer Elsayed}\\
  Computer Science and Engineering Department, Qatar University
  \\
  \texttt{\{sa1800633, ma1403845, se1403101, telsayed\}@qu.edu.qa}
}

\begin{document}
\maketitle
\begin{abstract}
Automated Essay Scoring (AES) faces significant challenges in cross-prompt settings, where models must generalize to unseen writing prompts. To address this limitation, we propose \sys, a meta-learning framework that leverages prototypical networks to learn transferable representations across different writing prompts. Across three diverse datasets (ELLIPSE and ASAP (English), and LAILA (Arabic)), \sys{} achieves state-of-the-art performance on ELLIPSE and LAILA, outperforming strong baselines by 8.5 and 3 points in QWK, respectively. On ASAP, where prompts exhibit heterogeneous score ranges, \sys{} yields improvements on several traits, highlighting the strengths of our approach in unified scoring settings. Overall, our results demonstrate the potential of meta-learning for building robust cross-prompt AES systems.
\end{abstract}

\section{Introduction}

Automated Essay Scoring (AES) has been an active research area for about six decades \cite{page1966imminence}. AES systems assess the quality of essays by providing holistic scores, trait-specific scores, or both
, offering scalable alternatives to costly manual grading in large-scale assessments \cite{burstein2013handbook}.

Current AES research follows two main paradigms: \textit{prompt-specific} and \textit{cross-prompt}.\footnote{A prompt is the description of a writing task.}
Prompt-specific AES trains and tests models on essays from the same prompt, achieving strong performance but requiring substantial labeled data \cite{kumar-etal-2022-many}.
In contrast, cross-prompt AES trains a model on a set of source prompts and tests it on \emph{unseen} target prompts \cite{ridley2021automated}. This setup is more practical, but struggles with performance due to prompt variability. 


Despite ongoing efforts, cross-prompt AES remains under-explored, with generalization still a major challenge. To contribute towards addressing this gap, we introduce \sys, a \textbf{M}et\textbf{A}-learning framework for \textbf{P}rototypica\textbf{L} cross-prompt \textbf{E}ssay scoring. Unlike prior optimization-based meta-learning approaches for AES \cite{chen-li-2024-plaes, wang2025making}, which treat each prompt as a separate meta-task and thus limit task diversity, \sys{} adopts a non-parametric approach based on prototypical networks \cite{snell2017prototypical} and proposes a more flexible meta-task definition that extends beyond prompts to include traits and score distributions. This design increases task heterogeneity and better leverages meta-learning’s strength in generalization, as suggested by prior work on heterogeneous task setups \cite{iwata2020meta, van2021multilingual}.

Inspired by the work of \citet{wu2021prototransformer}, we explore reformulating 
multiclass scoring as a series of binary classification tasks, which enhances task diversity during meta-training and improves the model’s generalizability. Additionally, we introduce a gating mechanism that incorporates contextual information from the prompt, trait-specific rubrics, and engineered features previously shown to enhance cross-prompt performance. 

In this work, we test \sys{} over essays of two different languages, English and Arabic, to assess its robustness. 
Experimental results show that \sys{} achieves SOTA performance on both ELLIPSE English dataset \cite{crossley2023english} and 
LAILA Arabic dataset \cite{LAILA2026}. On ASAP English dataset \cite{mathias2018asap++}, which presents additional challenges due to its varying score ranges across prompts, \sys{} achieves improvements on traits with unified score ranges.
This indicates that the proposed framework is a promising direction for building robust cross-prompt AES models. 

The main contribution of this work is four-fold:
\begin{enumerate}
    \item We introduce \sys{}, a \textit{novel} meta-learning framework for cross-prompt AES.
    \item We evaluate \sys{} on both English and Arabic essays to demonstrate its effectiveness and robustness across languages.
    \item \sys{} achieves SOTA performance on ELLIPSE and LAILA, and exhibits a competitive trait-level performance on ASAP.
    \item We release our implementation to support replication for future AES research.\footnote{\url{https://github.com/salbatarni/ACL2026_MAPLE}}  
\end{enumerate}

The remainder of this paper is organized as follows. We review related work in \S\ref{sec:related-work}, provide meta-learning background in \S\ref{sec:back-ground}, and introduce \sys{} in \S\ref{sec:maple}. The experimental setup is described in \S\ref{sec:exp-setup}, followed by results and discussion in \S\ref{sec:exp-eval}. Finally, we conclude and suggest future directions in \S\ref{sec:conclusion}.

\section{Related Work}\label{sec:related-work}
In this section, we review cross-prompt and Arabic AES, along with the main meta-learning frameworks and their adoption in existing AES systems. 

\subsection{Cross-prompt AES}
The limited feedback provided by cross-prompt holistic scoring \cite{Domain-Adaptive-2020,ridley2020prompt,pairwise-holistic-2025} motivated shifting towards trait scoring \cite{ridley2021automated}.
Recent advances include ProTACT \cite{do-etal-2023-prompt}, which combines prompt-aware essay representations with extracted features using POS-embedding-based neural model. \citet{chen-li-2023-pmaes} use contrastive learning to align source-target representations.
\citet{li-ng-2024-conundrums} adopt multi-task learning with a purely feature-based approach. 
\citet{EPCTS-2025} integrates 
syntactic embeddings with an LLM to measure essays' relevance. \citet{eltanbouly2025trates} scores essays using LLM-extracted and hand-crafted features. 
\citet{chen-etal-2025-mixture-ordered} proposed MOOSE, a Mixture-of-Experts (MoE) architecture, integrating prompt and essay information using experts for overall and relative quality, and essay-prompt relevance, achieving SOTA performance.

\subsection{Arabic AES}
Research on Arabic AES has recently gained momentum with the release of new publicly available datasets, including QAES \cite{bashendy-etal-2024-qaes} and TAQEEM 2025 \cite{bashendy-etal-2025-taqeem}. \citet{sayed2025feature} pioneered cross-prompt Arabic AES by introducing a comprehensive feature set and evaluating various AES models. 
TAQEEM 2025 shared task introduced Arabic cross-prompt holistic and trait scoring tasks, with a multitask AraBERT baseline \cite{bashendy-etal-2025-taqeem}. The top-performing systems employed GPT-4o with few-shot prompting \cite{almarwani-etal-2025-taibah}, and GPT-4.1 with 10-shot chain-of-thought prompting \cite{alnajjar-etal-2025-arxhyoka}.
Most recently, LAILA dataset \cite{LAILA2026} was introduced with 
7,859 essays annotated with seven traits and a holistic score, and was benchmarked in a cross-prompt setup using feature-based models, adapted English SOTA baselines, and LLM-based approaches.

\subsection{Meta-learning for AES}
In the AES context, PLAES \cite{chen-li-2024-plaes}, a framework for cross-prompt AES, was proposed employing MAML \cite{finn2017model} to capture general knowledge across prompts, and a level-aware contrastive learning to distinguish the different essays' quality. In addition, \citet{wang2025making} proposed MLCAES, a meta-learning framework for holistic cross-prompt AES that guides model generalization toward target prompt distributions using a distribution-guided meta-learner selection mechanism. 
Both approaches fall under optimization-based methods, where the model is trained to directly optimize its parameters for rapid adaptation.
\citet{zeng2023generalizable} used prototypical networks, a non-parametric method, for cross-prompt \textit{short-answer} scoring. 
While we also adopt prototypical networks, we address \textit{essay} scoring, which differs 
from short-answer scoring in both length and trait complexity.
Moreover, their approach relies on few-shot labels from the target prompt, whereas we assume \textit{completely unseen target prompts}. In addition, we explore multiple task-construction strategies (\S\ref{sec:task-formulation}) rather than a single-task setup, jointly predict multiple traits instead of a single holistic score, and retain the original prototypical loss while incorporating hand-crafted features, all of which distinguish our approach.

Despite these advances, existing meta-learning approaches for AES tend to model prompts as homogeneous tasks, conflicting with meta-learning’s strength in heterogeneous settings \cite{iwata2020meta, van2021multilingual}. This limits tasks to available prompts, reducing generalization. We follow \cite{wu2021prototransformer} by exploring reframing the problem as binary classification to increase task diversity and improve model adaptability and effectiveness.

\section{Background: Meta-learning}\label{sec:back-ground}
Meta-Learning is a few-shot learning paradigm that enables rapid adaptation from task-independent to task-specific spaces \cite{finn2017model},
with success across different domains \cite{triantafillou2019meta, tarunesh2021meta, wu2021prototransformer}.

Meta-learning consists of \emph{two phases}: \emph{meta-training} and \emph{meta-testing}. During meta-training, a model is exposed to a diverse set of tasks, each with labeled support and query sets, to learn adaptable representations that generalize across tasks. In meta-testing, the model supposedly adapts to a new task, leveraging labeled support set to predict labels for unlabeled query set. In our work, we employ \emph{Prototypical Networks} \cite{snell2017prototypical}, a meta-learning framework that learns a shared embedding space for classification problems, where \emph{class prototypes} are computed based on corresponding labeled support set examples. These prototypes enable classification of unlabeled query samples by proximity in the embedding space. 

\paragraph{Meta-training} During meta-training, the model is trained on a set of classification tasks $\mathcal{T}_{tr}=\{T_i\}$. Each task $T_i$ consists of a pair of labeled subsets: a \emph{support} set $S_i$ and a \emph{query} set $Q_i$. The support set contains $k$ shots (examples) from every class $c\in C_i$, where $C_i$ is the class set of 
$T_i$. All examples are encoded using an encoder or learner $f_\theta$, which maps an input $x$ into an embedding space $f_\theta(x) \in \mathbb{R}^d$. 

\begin{figure}
    \centering
    \includegraphics[width=0.9\linewidth]{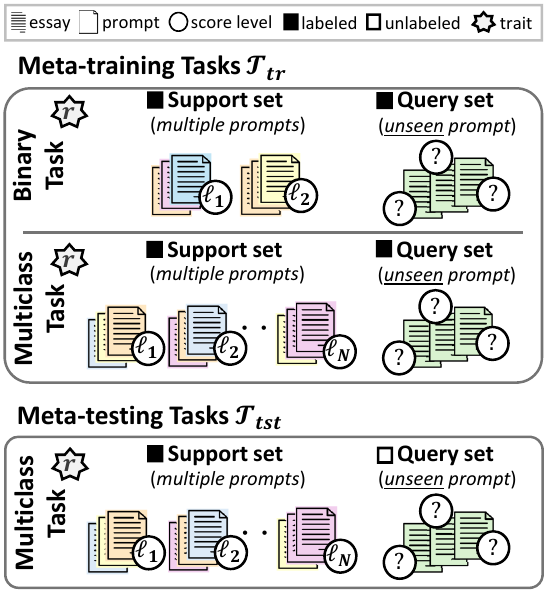}
    \caption{\sys{} task generation. In meta-training, we explore two settings, binary and multiclass classification. Each sampled task includes a support set (for $C_i$ computation) and a query set (for evaluation/learner-update). In meta-testing, the task is multiclass, where the support set includes all training data and the query set corresponds to an unseen prompt.}
    \label{fig:tasks}
\end{figure}

Formally, the support set $S_i$ is defined as $\{(x_j^s,y_j^s)\}$, where $x_j^s$ is an example and $y_j^s$ is its corresponding class. Similarly, the query set $Q_i$ is a set of labeled examples $\{(x_j^q, y_j^q)\}$; 
the model uses the support set to predict the classes of the query set by learning an embedding space where each class $c$ is represented by a prototype $c^\star$, computed as the centroid of embeddings of its $k$-shot examples:  
\begin{equation}\label{equ:centroids}
    c^\star=\frac{1}{k}\sum_{y_j^s = c}{f_\theta(x_j^s)} 
\end{equation}
For each query example $x_j^q$, the learner computes the distance $D$, using 
a distance function $\phi$, between 
$f_\theta(x_j^q)$ and the class prototypes:

\begin{equation}\label{equ:distance}
    D(x_j^q, c) = \phi(f_\theta(x_j^q), c^\star), c \in C_i
\end{equation}
The final prediction $c_j^q$ 
is assigned to the class closest to the query sample:
\begin{equation}\label{equ:prediction}
    c_j^q=\underset{c \in C_i}{\arg\max}\ D(x_j^q, c)
\end{equation}
Finally, the model is updated based on its performance on the query set, with the loss function $\mathcal{L}$:
\begin{equation}\label{equ:loss}
    \mathcal{L}(x_j^q) = -\log \frac{\exp\left({D(x_j^q, y_j^q)}\right)}{\sum_{c \in C_i} \exp \left(D(x_j^q, c) \right)}
\end{equation}
where $y_j^q$ is the class label of $x_j^q$.

\paragraph{Meta-testing} After training the learner $f_\theta$, the model is ready to handle new, unseen tasks $\mathcal{T}_{tst}=\{T_i\}$. Similar to meta-training, each task ${T}_i$ consists of a support set $S_i$ and a query set $Q_i$. The key difference, however, is that the query set $Q_i$ in meta-testing is \emph{unlabeled}. The learner $f_\theta$ quickly adapts to each new task ${T}_i$ using the \textit{labeled} support set $S_i$, then predicts the labels for the examples in the query set $Q_i$, following the same methodology as in meta-training. This formulation avoids reliance on few-shot parameter adaptation from prompt-specific data and supports robust generalization to completely unseen prompts.

\section{\sys{} for Cross-Prompt AES}\label{sec:maple}

\begin{figure}[t]
  \centering
  \begin{minipage}[b]{0.4\textwidth}
    \centering
    \includegraphics[width=\textwidth]{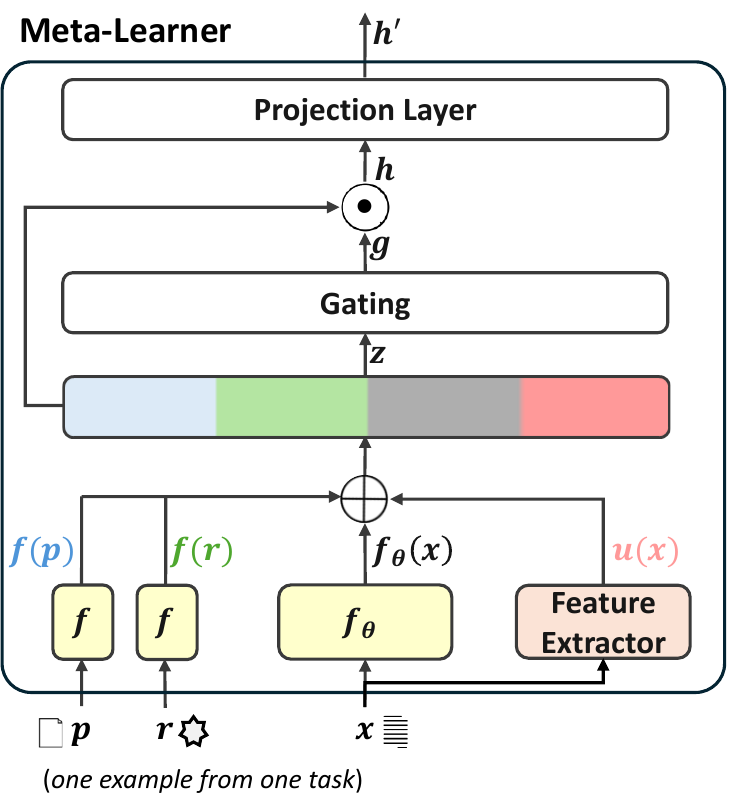}
    \caption*{(a) Meta-learner architecture. The input consists of an essay $x$, trait rubric $r$, and prompt $p$. The output is the essay representation $h'$.}
    \label{fig:sub-a}
  \end{minipage}
  
  \vspace{1em}
  \begin{minipage}[b]{0.4\textwidth}
    \centering
    \includegraphics[width=\textwidth]{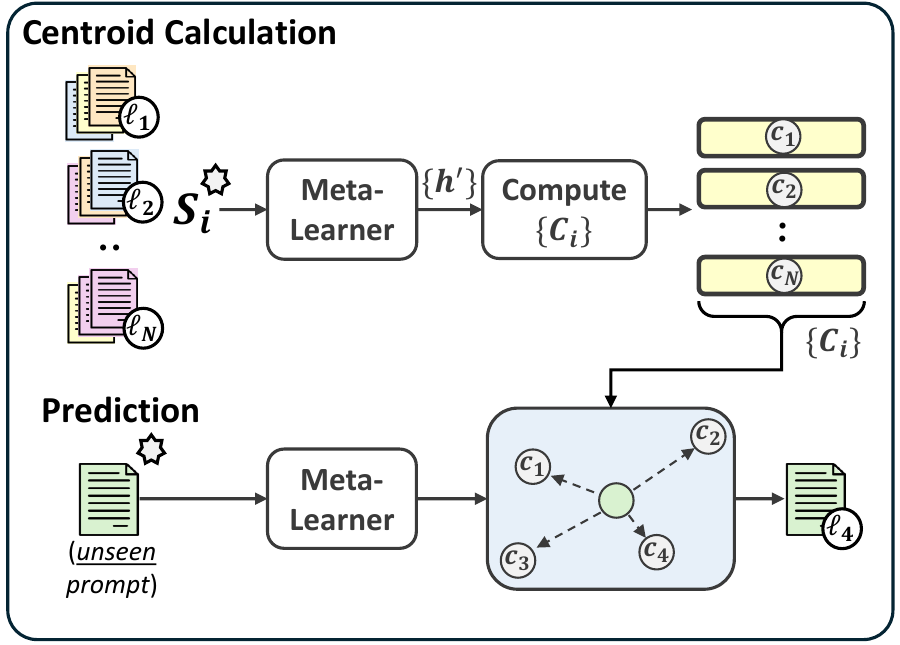}
    \caption*{(b) The prediction process involves two steps. The meta-learner first encodes the support-set for centroids $C_i$ computation. Then, the unseen essay representation is compared with these centroids to predict the score $l$.}
  \end{minipage}

  \caption{Overview of \sys{} showing (a) the meta-learner architecture and (b) the prediction process.}
  \label{fig:maple}
\end{figure}

In this section, we describe \sys{} by introducing our meta-learning task formulation for cross-prompt AES, outlining our different training strategies (Figure~\ref{fig:tasks}), and presenting our model architecture and prediction process (Figure~\ref{fig:maple}).

\subsection{Meta-training Task Formulation} \label{sec:task-formulation}
Cross-prompt AES aims to train a model on essays from a set of \emph{source} prompts $P^s = \{p_i^s\}$ and test its generalization capability on essays from an unseen \emph{target} prompt $p^t$. A direct adaptation of cross-prompt AES into the meta-learning framework is to frame each prompt ${p^{s}_i}$ as a distinct task ${T}$. However, this approach presents a significant challenge: 
the number of unique meta-training tasks is limited by the source prompts, reducing task diversity which is essential for effective meta-learning.

To address this, we propose two training formulations: \begin{inparaenum}[i)] \item \emph{binary} classification, where the number of tasks is bounded by the combination of prompts, traits, and score levels; and
\item \emph{multiclass} classification, where the number of tasks is bounded by the combination of prompts and traits.\footnote{We note that we follow \citet{wu2021prototransformer} by excluding tasks that have less than $k+m$ samples, where $k$ and $m$ denote the numbers of samples in the support set and query set, respectively.} These settings are shown in the meta-training tasks in Figure~\ref{fig:tasks}.
\end{inparaenum}

\paragraph{Binary classification} 
Inspired by \citet{wu2021prototransformer}, each meta-training task $T$ in this setup is represented as a tuple $(p, r, l^\star)$, where $p$ is a prompt from which essays are sampled, $r$ is a writing trait, and $l^\star$ is a score level. For each sampled task, we construct two classes: 
a positive class
$(p, r, l=l^\star)$ and a negative class 
$(p, r, \overline{l} \neq l^\star)$
from which we sample $k$ essays each, substantially increasing task diversity for potentially improving meta-learning performance \cite{iwata2020meta}.

\paragraph{Multiclass classification} In this setup, each meta-training task $T$ is 
represented as a tuple $(p, r)$, where $p$ is a prompt and $r$ is a writing trait. For each sampled task, we draw $k$ examples from \emph{each} score level of trait $r$, representing a class per level. While this reduces the number of tasks compared to the binary formulation, it better mirrors the actual scoring process, which is inherently a multiclass classification task.

In both setups, every task is constructed as a cross-prompt task, i.e., the support set and query set \emph{always} come from different prompts, as illustrated in Figure~\ref{fig:tasks}. Specifically, we first sample one prompt for the query set, and then select a different prompt (or prompts) for the support set. In fact, we experiment with two variants for constructing the support set: \textbf{one-prompt} (denoted as \textbf{1P}), where the support set is drawn from one single prompt different from the query prompt, and \textbf{multiple prompts} (denoted as \textbf{mP}), where the support set is drawn from all available prompts except the query prompt.

\subsection{Meta-testing}

After training the learner $f_\theta$, the model is evaluated on new tasks $\mathcal{T}_{tst} = \{{T_i}\}$ constructed from \emph{unseen} prompts. Each task $T_i$ is defined by a trait $r$ and a prompt $p_{tst}$, formulated as an $N$-way classification problem, where $N$ is the number of score levels of trait $r$. To satisfy the cross-prompt condition, the support set $S_i$ is sampled from training essays written for different prompts other than $p_{tst}$, while the query set $Q_i$ consists of essays from $p_{tst}$.

For each trait $r$, prototypes are constructed from the support set by averaging the embeddings of essays that share the same score level, as defined in Equation~\ref{equ:centroids}. In this case, $k$ corresponds to all training essays with score $l$ for trait $r$. Each query essay $x_j^q$ is then scored by assigning it to the nearest prototype centroid, following Equation~\ref{equ:prediction}.

\subsection{Incorporating More Context}

We anticipate a potential bottleneck in the above approach; although the centroids are computed for each trait separately, the representations of the essays are the same for different traits. 
Furthermore, having essays from different prompts within the same meta-learning task necessitates providing richer contextual information about those prompts.
Accordingly, we include the \textbf{prompt and rubric} texts as additional inputs to the learner. 

Figure \ref{fig:maple}(a) shows the architecture of the meta-learner model. $f(p)$ and $f(r)$ denote the prompt and rubric representations, respectively, obtained by the \emph{pretrained} encoder $f$. $f_\theta(x)$ is the learned essay representation by the learner $f_\theta$. To effectively combine these sources of information, we concatenate them into a joint vector, $z = [f_\theta(x); f(p); f(r)]$, and introduce a gating mechanism that allows the model to selectively weight each component in a task-dependent manner.\footnote{Preliminary experiments with existing attention-based fusion and GLU showed lower performance.} This adaptive fusion 
ensures that context from prompts and rubrics is emphasized when relevant for each task. The element-wise gates are computed as:
\begin{equation}
g = \sigma(W_z z),
\end{equation}
where $W_z \in \mathbb{R}^{3d \times 3d}$ is a learnable parameter matrix and $\sigma(\cdot)$ is the sigmoid function. The gated representation $h$ is obtained as:
\begin{equation}
h = z \odot g,
\end{equation}
where $\odot$ denotes element-wise multiplication.
Finally, $h$ is passed through a projection layer:
\begin{equation}
h' = W_2 \, ReLU \!\left(\mathcal{O}\!\left(W_1 h + b_1\right)\right) + b_2,
\end{equation}
where $W_1 \in \mathbb{R}^{3d \times 3d}, W_2 \in \mathbb{R}^{3d \times d}$,  $b_1$, and $b_2$ are learnable parameters, and $\mathcal{O}(\cdot)$ denotes dropout with rate 0.5. The output $h' \in\mathbb{R}^d$ serves as the final representation for scoring.

Moreover, engineered features have proven useful, particularly in cross-prompt setups \cite{do-etal-2023-prompt, sayed2025feature}. Hence, we explore incorporating a set of engineered features, $u(x)\in\mathbb{R}^{d_u}$, into the essay representation. Specifically, we extend $z$ to be 
$[f_\theta(x); f(p); f(r); u(x)]$
and apply the same gating and projection steps to obtain $h'$. The prediction process is outlined in Figure~\ref{fig:maple}(b).

\section{Experimental Setup}\label{sec:exp-setup}
This section outlines the experimental setup, covering encoder selection, training and meta-learning specifications, hyper-parameters, datasets, evaluation metric, and baselines. 

\subsection{Hand-crafted Features}
For ELLIPSE and ASAP, we use the 86 feature-set proposed by \citet{ridley2020prompt}, which covers length-based, readability, text complexity, text variation, and sentiment features. 
For Arabic, we use the 816 features introduced by \citet{sayed2025feature}, including surface, readability, lexical, semantic, and syntactic aspects.

\subsection{Encoder Selection}
Encoder model selection was based on GPU compatibility and performance,
leading us to choose models under 200M parameters: AraBERTv2\footnote{\url{https://huggingface.co/aubmindlab/bert-base-arabertv2}}
for Arabic and RoBERTa\footnote{\url{https://huggingface.co/FacebookAI/roberta-base}} for English.

\subsection{Training and Hyperparameters}
All experiments were conducted using 4 NVIDIA A10-24Q GPUs with PyTorch mixed precision training and Adam optimizer. The number of shots in meta-training was fixed to 5 for both the support ($k$) and query ($m$) sets. For the multiclass classification setup, the maximum number of classes was set to 5 to align with computational constraints. The model was trained for 30K meta-training tasks with a batch size of 12 (the maximum our hardware resources could accommodate). The best-performing model on the dev set was selected for evaluation. 

During the meta-training setup selection phase, the learning rate and number of learnable encoder layers were initially fixed at $1e^{-5}$ and 3, respectively. Before training the final model for testing, these hyperparameters were tuned over the following values: for AraBERT, learning rate $\in \{1e^{-5}, 5e^{-6}, 1e^{-6}\}$ and number of tunable layers $\in \{3, 6, 9\}$; for RoBERTa, learning rate $\in \{1e^{-5}, 3e^{-5}, 5e^{-5}\}$ and number of tunable layers $\in \{3, 6, 9\}$. 
The best hyper-parameters for each dataset are reported in Appendix \ref{sec:appendix-best-hyperparameters}.


\subsection{Datasets and Evaluation}\label{sec:datasets}

\paragraph{English Data} For experimenting on English data, we use two datasets: ELLIPSE \cite{crossley2023english} and ASAP\footnote{\url{https://www.kaggle.com/c/asap-aes}}/ASAP++ ~\cite{mathias2018asap++}. 
ELLIPSE serves as our main English dataset, as it provides a consistent score range across prompts, making it well-suited to demonstrate the strengths of \sys{}. ASAP is primarily used to compare \sys{} to existing baselines; however, its varying score ranges across prompts make some traits more challenging, as certain score levels are unrepresented in the support set.

ELLIPSE comprises about 6.5k essays written by English Learners for 44 distinct prompts. The essays, written by students in grades 8-12, are assessed across 7 traits: cohesion (COH), syntax (SYN), vocabulary (VOC), phraseology (PHR), grammar (GRM), conventions (CNV), and holistically (HOL), with an average length of 427 words. Each trait is scored using a standardized rubric on a 1-5 scale, with 0.5-point increments. ASAP dataset contains approximately 13k essays across 8 prompts, where trait-level annotations are available for prompts 7 and 8. ASAP++ extends ASAP by providing trait annotations for prompts 1–6. The evaluated traits include Content (CNT), Organization (ORG), Word Choice (WC), Sentence Fluency (SF), Conventions (CNV), Prompt Adherence (PA), Language (LNG), and Narrativity (NAR). Since score ranges vary across prompts in ASAP, we shift all scores to start from 0 to align them across prompts.
ASAP per-prompt traits and score ranges are described in Appendix \ref{sec:asap-stat}.\footnote{We refer to ASAP/ASAP++ as ASAP hereafter.} Although ASAP dataset includes holistic scores, the score ranges vary substantially across prompts. Thus, we exclude holistic scoring and focus on trait-level evaluation, leaving this issue for future work.


\paragraph{Arabic Data} 
For Arabic, we used \textbf{LAILA} dataset \cite{LAILA2026}, comprising 7,859 essays written by native high school students under test-like conditions across 8 different prompts. Each essay is annotated across 7 traits: Relevance (REL), Organization (ORG), Vocabulary (VOC), Style (STY), Development (DEV), Mechanics (MEC), and Grammar (GRM), in addition to a Holistic (HOL) score computed as the sum of all trait scores, with an average length of 171 words. 
All traits are assessed on a scale of 0-5, except REL on a scale of 0-2, using 1-point increments. 
 
\paragraph{Data Splits} We employed different cross-validation strategies for the English and Arabic datasets. 
For ELLIPSE, we followed the 11-fold cross-validation splits proposed by \citet{eltanbouly2025trates}, with 40 prompts for training and 4 unseen prompts for testing. Within the training set, each prompt is partitioned into training and development subsets using an 80/20 split. 
For ASAP, we adopt the 8-fold leave-one-prompt-out cross-validation setup of \citet{ridley2020prompt}, where for each fold, seven prompts are used for training and development, and one unseen prompt is reserved for testing.
For LAILA, we use the public cross-prompt splits provided by the dataset authors,\footnote{\url{https://gitlab.com/bigirqu/laila}} following an 8-fold leave-one-prompt-out setup, with 5 training, 2 development, and 1 test prompt for each fold.

\paragraph{Evaluation Metric} To assess model performance, we employ Quadratic Weighted Kappa (QWK)~\cite{cohen1968weighted}, a widely adopted metric in AES that quantifies agreement between human annotators and system predictions.

\subsection{Baselines}
We compare our proposed approach with SOTA baseline models for both English and Arabic AES.
\paragraph{English Baselines} 
On ELLIPSE, we compare with TRATES \cite{eltanbouly2025trates} (the SOTA method on ELLIPSE) and MOOSE \cite{chen-etal-2025-mixture-ordered} (the SOTA method on ASAP), which we ran on ELLIPSE using their publicly available code.\footnote{\url{https://github.com/antslabtw/MOOSE-AES}}

On ASAP, we compare with MOOSE and EPCTS \cite{EPCTS-2025}, the current two best performing models on ASAP. 
Upon examination of the released code,\footnotemark[\value{footnote}] we note that the reported results of MOOSE on ASAP (Table 3 in \cite{chen-etal-2025-mixture-ordered}) were based on tuning on the \emph{test} set, rather than the \emph{dev} set. We confirmed that by running the code and managed to almost reproduce the reported results.\footnote{We obtained a score of 0.635 compared to 0.640 (avg w/o overall) reported in the paper.} However, with properly tuning on the dev set,\footnote{Using the same set of the fixed hyperparameters outlined in \cite{chen-etal-2025-mixture-ordered}.} we obtained an average score of 0.538. Table \ref{tab:test-results} reports both the original paper's test-optimized results (which are
not comparable to the others, and marked by *) and also the dev-optimized results for fair and transparent comparison.


\paragraph{Arabic Baselines}
We select the best performing model on LAILA under the cross-prompt setup, specifically MOOSE, where \citet{LAILA2026} replaces BERT with AraBERT and incorporates features proposed by \citet{sayed2025feature}. Additionally, we compare with XGB, the second-best model, which also uses the same set of features.


\section{Experimental Evaluation}\label{sec:exp-eval}

We aim to address the following research questions in the context of cross-prompt AES:
\begin{inparaenum}[({RQ}1)]
\textbf{\item} Which meta-training formulation, multiclass or binary classification, yields the best performance?
\textbf{\item} How does incorporating prompt and rubric information influence performance?
\textbf{\item} What is the impact of integrating hand-crafted features on the performance?
\textbf{\item} How does \sys{} compare to baseline SOTA models for English and Arabic AES?
\end{inparaenum}

This section answers the research questions and discusses the corresponding results. We report performance results on dev sets as we navigate through the best configurations of our framework (the first three RQs). Finally, we compare the best setup with SOTA on test sets (RQ4). 

\subsection{Meta-training Setups (RQ1)}


\begin{table}[t]
\centering
\begin{tabular}{llc} 
\hline
Dataset                                   & Setup         & Avg              \\ 
\hline
\multirow{6}{*}{ELLIPSE}                  & Binary-1P     & 0.548            \\
                                          & Binary-mP     & 0.574            \\
                                          & Multiclass-1P & 0.578            \\
                                          & Multiclass-mP & \textbf{0.581}   \\ 
\cline{2-3}
                                          & +PR           & 0.587$^\bullet$  \\
                                          & +PR +Features & 0.592$^\bullet$  \\ 
\hline\hline
                                          & Setup         & Avg              \\ 
\cline{2-3}
\multicolumn{1}{c}{\multirow{6}{*}{ASAP}} & Binary-1P     & 0.469            \\
\multicolumn{1}{c}{}                      & Binary-mP     & 0.474            \\
\multicolumn{1}{c}{}                      & Mutliclass-1P & \textbf{0.622}   \\
\multicolumn{1}{c}{}                      & Multiclass-mP & 0.603            \\ 
\cline{2-3}
\multicolumn{1}{c}{}                      & +PR           & 0.639$^\bullet$  \\
\multicolumn{1}{c}{}                      & +PR +Features & 0.645$^\bullet$  \\ 
\hline\hline
                                          & Setup         & Avg              \\ 
\cline{2-3}
\multirow{6}{*}{LAILA}                    & Binary-1P     & 0.599            \\
                                          & Binary-mP     & \textbf{0.601}   \\
                                          & Multiclass-1P & 0.596            \\
                                          & Multiclass-mP & 0.588            \\ 
\cline{2-3}
                                          & +PR           & 0.614$^\bullet$  \\
                                          & +PR +Features & 0.631$^\bullet$  \\
\hline
\end{tabular}
\caption{Average performance of the meta-training setups in QWK on the \underline{dev sets}. Bold indicates the best-performing setup, and $^\bullet$ marks improvements over the best setup. PR indicates prompt and rubric.}
\label{tab:dev-results}
\end{table}

RQ1 examines the different meta-learning setups we proposed in \S\ref{sec:task-formulation}. We consider four setups: \textbf{binary-1P}, \textbf{binary-mP}, \textbf{multiclass-1P}, and \textbf{multiclass-mP}, where binary/multiclass indicates the task classification type, and 1P/mP specifies whether the support set is sampled from a single or multiple prompts.
Datasets' average performance on the dev sets is reported in Table \ref{tab:dev-results} (detailed per-trait results are shown in Table \ref{tab:dev-results-full} in Appendix \ref{sec:appendix-full}). 

On ELLIPSE, multiclass classification tasks outperform the binary tasks and are consistent across different support-set sampling strategies, achieving gains of 3 and 1 points for the \textbf{1P} and \textbf{mP} strategies, respectively. 
Over ASAP, multiclass tasks outperform binary tasks; however, the \textbf{1P} strategy performs better than \textbf{mP} by approximately 2 points.
On LAILA, in contrast, binary setups generally perform better. 
Although this differs from the inference scenario, it supports prior findings that more tasks improve performance \cite{wu2021prototransformer}.


The differences across datasets arise from the number of prompts and score granularity. 
LAILA has few prompts with coarse scores, making binary classification effective by reducing confusion and yielding stable prototypes. ASAP also has few prompts but uses heterogeneous score ranges, hence, aggregating across prompts leads to incompatible prototypes, favoring the 1P setup. In contrast, ELLIPSE has many prompts with fine-grained scores and limited samples per score, where multiclass classification better captures subtle distinctions. Based on these observations, we adopt the \textbf{Multiclass-mP}, \textbf{Mutliclass-1P}, and \textbf{Binary-mP} setups for subsequent experiments on ELLIPSE, ASAP, and LAILA, respectively.

\subsection{Effect of Adding Writing Context (RQ2)}


\begin{table*}
\centering
\begin{tabular}{llccccccccc} 
\hline
Dataset                  & Model  & COH            & SYN            & VOC            & PHR            & GRM            & CVN            & HOL            & Avg$^{-H}$        & Avg             \\ 
\hline
\multirow{3}{*}{ELLIPSE} & MOOSE  & 0.207          & 0.227          & 0.168          & 0.209          & 0.146          & 0.216          & 0.226          & 0.195          & 0.200           \\
                         & TRATES & 0.519          & 0.540          & 0.522          & 0.525          & 0.512          & 0.561          & -              & 0.530          & -               \\
                         & \sys{}   & \textbf{0.575} & \textbf{0.616} & \textbf{0.617} & \textbf{0.639} & \textbf{0.607} & \textbf{0.633} & \textbf{0.700} & \textbf{0.615} & \textbf{0.627}  \\ 
\hline\hline
Dataset                  & Model  & CNT            & ORG            & WC             & SF             & CNV            & PA             & LNG            & NAR            & Avg             \\ 
\hline
\multirow{4}{*}{ASAP}    & MOOSE*  & 0.651          & 0.652          & 0.634          & 0.643          & 0.604          & 0.649          & 0.624          & 0.665          & 0.640           \\
                         & MOOSE & 0.559          & 0.533          & 0.570          & 0.559          & 0.464          & 0.542          & 0.506          & 0.569          & 0.538           \\
                         & EPCTS  & \textbf{0.630}          & \textbf{0.606}          & \textbf{0.614}          & \textbf{0.617}          & \textbf{0.525}          & 0.630          & 0.613          & 0.647          & \textbf{0.610}           \\
                         & \sys{}   & 0.555          & 0.465          & 0.483          & 0.529          & 0.482          & \textbf{0.650} & \textbf{0.633} & \textbf{0.685} & 0.560           \\ 
\hline\hline
Dataset                  & Model  & REL            & ORG            & VOC            & STY            & DEV            & MEC            & GRA            & HOL            & Avg             \\ 
\hline
\multirow{3}{*}{LAILA}   & MOOSE  & \textbf{0.411} & 0.627          & 0.642          & 0.649          & 0.585          & 0.586          & 0.623          & 0.649          & 0.597           \\
                         & XGB    & 0.360          & 0.645          & 0.641          & 0.641          & 0.583          & 0.577          & 0.619          & 0.679          & 0.593           \\
                         & \sys{}   & 0.290          & \textbf{0.647} & \textbf{0.686} & \textbf{0.702} & \textbf{0.607} & \textbf{0.634} & \textbf{0.664} & \textbf{0.723} & \textbf{0.619}  \\
\hline
\end{tabular}
\caption{\sys{} performance in QWK on the \underline{test sets} compared to SOTA baselines. Avg$^{-H}$ denotes average performance without HOL scoring. \textbf{Bold} values indicate best performance per dataset per trait, excluding MOOSE*.}
\label{tab:test-results}
\end{table*}

We next investigate whether incorporating the rubric and prompt information improves performance by providing richer task context and clearer distinctions across traits and topics. 
Results in Table \ref{tab:dev-results} show that the performance improves over all datasets when prompt and rubric representations are included, but the improvement is more pronounced on LAILA and ASAP than ELLIPSE.

Examining the trait-level results, we note that prompt-independent traits benefit more from incorporating the rubric and prompt information than prompt-dependent traits. Additionally, larger gains are observed on traits with detailed rubrics and datasets with longer prompt texts. More detailed per-trait analyses for each dataset are provided in Appendix~\ref{sec:appendix-rq2}.


\subsection{Effect of Adding Features (RQ3)}
Given the performance boost gained by incorporating the contextual representations on all datasets, we next examine the effect of incorporating a feature-engineered vector into the essay representation. As shown in Table~\ref{tab:dev-results}, incorporating these features had a positive influence over all datasets, with an improvement of about 0.7 points on ASAP, 0.5 points on ELLIPSE, and 2 points on LAILA. 

Examining the trait-level results (detailed in Appendix~\ref{sec:appendix-rq3}), we note that improved traits emphasize structural, lexical, or surface-level aspects of writing, such as grammar and conventions on ELLIPSE, organization and content on ASAP, and organization and vocabulary on LAILA. These results suggest that the features capture aspects of clarity, correctness, and structure.


\subsection{\sys{} vs. SOTA (RQ4)}
Based on the above findings, 
we evaluate \sys{} on the test set of the three datasets using the configuration that incorporates the prompt, rubric, and feature information. Table~\ref{tab:test-results} compares \sys{} against SOTA baselines over the three datasets. 

\paragraph{ELLIPSE}
\sys{} outperforms the best baseline model by 8.5 points on ELLIPSE (on average across all traits but HOL, since TRATES, the best baseline, focuses solely on trait-level scoring). 
The improvement is pronounced on \emph{all traits} by 5-10 points.
These results establish a new important benchmark for ELLIPSE, a dataset that has been underutilized despite addressing limitations in ASAP \cite{li-ng-2024-automated}.

\paragraph{ASAP}
On average, \sys{} outperforms MOOSE but lags behind EPCTS. This comparatively-lower performance is expected, since ASAP prompts have \emph{heterogeneous} score ranges (see Table~\ref{tab:asap-dataset} in Appendix \ref{sec:asap-stat}). For example, prompt 8 has 11 score levels, whereas other prompts have a maximum of 6, leaving 5 score levels unrepresented in the support set.\footnote{We leave addressing this issue to future work.} Nevertheless, \sys{} demonstrates strong performance on traits that have \emph{unified} score ranges, specifically PA, LNG, and NAR, achieving 2-4 points improvements over SOTA models and setting new SOTA performance for these traits. Detailed ASAP results, including per-prompt comparisons with baselines, as well as per-trait and per-prompt breakdowns, are provided in Appendix \ref{sec:full-results}.

\paragraph{LAILA}
\sys{} outperforms the best baseline models by 3 points on LAILA, achieving an average gain of 4 points across all traits except REL. The improvements are most pronounced on HOL with a 7-point gain, and on STY and MEC with 5-point gains each. For transparency, we report per-prompt and per-trait results in Appendix \ref{sec:full-results}.

\textbf{Overall}, \sys{} exhibits SOTA performance when score ranges are unified, as seen in ELLIPSE, LAILA, and the unified-range traits in ASAP, demonstrating its effectiveness in cross-prompt AES. Moreover, these findings underscore the importance of evaluating AES models across diverse datasets to obtain a more comprehensive assessment of their generalization ability.

\section{Conclusion and Future Work}\label{sec:conclusion}
In this work, we propose \sys, a meta-learning framework based on prototypical networks for cross-prompt AES that 
integrates prompt, rubric, and feature representations with essay embeddings to enhance generalization across prompts and traits. Through systematic experiments on three English and Arabic datasets, 
results showed that \sys{} achieves SOTA performance on two datasets, outperforming strong baselines by an impressive 8.5 points on ELLIPSE and up to 3 points on LAILA, demonstrating the potential of meta-learning for building adaptable, cross-prompt AES systems. 
ASAP dataset was more challenging due to its varying score ranges; however, for traits unaffected by this issue, improvements reached up to 4 points.
For future work, we plan to address the score range variability, explore auxiliary tasks to further enhance model generalizability, and extend the approach to multilingual settings by incorporating a language dimension into task definitions.

\section*{Acknowledgments}
This work was made possible by NPRP grant\# NPRP14S-0402-210127 from the Qatar Research Development and Innovation (QRDI) Council. The statements made herein are solely the responsibility of the authors.
 
\section*{Limitations}
While \sys{} demonstrates strong performance in cross-prompt AES, several limitations exist. 

First, 
we experimented with a single encoder model for each language, therefore, the effect of varying pre-trained models on \sys{}'s performance remains an open question. 

Second, we fixed the size of the support and query sets to 5 examples during meta-training for computational efficiency, though tuning this hyperparameter could improve the performance. 

Third, one limitation becomes evident in datasets with different score ranges across the prompts (e.g., ASAP dataset), as \sys{}, a classification-based approach, does not inherently handle differences in score ranges. Unlike regression models, where scores can be scaled to a unified range during training and rescaled to their original ranges during inference, such scaling is not intuitive in the classification setting. This is because score discretization requires rounding, which can introduce mismatches in the mapping of scores across prompts with different score ranges and when scaling the scores back during evaluation. 
Nevertheless, \sys{} demonstrates superior performance on datasets and prompts with unified score ranges. Adapting \sys{} to prompts or traits with varying score ranges remains open for future work.

Finally, although we explored the combined effect of prompt and rubric information, we did not isolate their individual contributions. A more granular investigation of each component could provide deeper insights into their specific roles in enhancing model performance.


\bibliography{references}

\appendix

\section{ASAP Statistics}\label{sec:asap-stat}

Table \ref{tab:asap-dataset} presents the statistics for ASAP dataset, including the traits associated with each prompt and their corresponding score ranges. We include these details primarily to highlight the variability in score ranges across prompts in ASAP dataset.

\section{\sys{} Hyperparameters}
\label{sec:appendix-best-hyperparameters}
We report the best hyperparameters obtained after tuning the learning rate and the number of trainable layers for each dataset in Table~\ref{tab:best-hyperparameters}.

\begin{table}[h]
\centering
\begin{tblr}{
colsep=1.5pt,
rowsep=1pt,
  column{2} = {c},
  column{3} = {c},
  hline{1-2,5} = {-}{},
}
Dataset & Learning rate & Trainable layers \\
ASAP    & $1e^{-5}$          & 3                \\
ELLIPSE & $1e^{-5}$          & 3                \\
LAILA   & $1e^{-5}$          & 9                
\end{tblr}
\caption{Best hyperparameters for each dataset.}
\label{tab:best-hyperparameters}
\end{table}

\begin{table*}[htp]
\centering

\begin{tblr}{
  row{even} = {c},
  rowsep=0pt,
  row{3} = {c},
  row{5} = {c},
  row{7} = {c},
  row{9} = {c},
  cell{1}{2} = {c},
  cell{1}{3} = {c},
  cell{1}{4} = {c},
  cell{1}{5} = {c},
  cell{1}{6} = {c},
  cell{1}{7} = {c},
  cell{1}{8} = {c},
  cell{1}{9} = {c},
  cell{1}{10} = {c},
  cell{1}{11} = {c},
  cell{1}{12} = {c},
  vline{2,5} = {-}{},
  hline{1-2,10} = {-}{},
}
Prompt & Scores & Ave Length & Essays & CNT                   & ORG              & WC              & SF          & CNV              & PA          & LNG                  & NAR                                 \\
P1    & 1 - 6         & 350        & 1783          &  $\checkmark$  &  $\checkmark$  &  $\checkmark$  &  $\checkmark$  &  $\checkmark$  &                           &                           &                                             \\
P2    & 1 - 6         & 350        & 1800          &  $\checkmark$  &  $\checkmark$  &  $\checkmark$  &  $\checkmark$  &  $\checkmark$  &                           &                           &                                                     \\
P3    & 0 - 3         & 100        & 1726          & $\checkmark$   &                           &                           &                           &                           &  $\checkmark$  &  $\checkmark$  &  $\checkmark$                             \\
P4    & 0 - 3         & 100        & 1772          &  $\checkmark$  &                           &                           &                           &                           & $\checkmark$ & $\checkmark$ & $\checkmark$                            \\
P5    & 0 - 4         & 125        & 1805          & $\checkmark$ &                           &                           &                           &                           & $\checkmark$ & $\checkmark$ & $\checkmark$                            \\
P6    & 0 - 4         & 150        & 1800          & $\checkmark$ &                           &                           &                           &                           & $\checkmark$ & $\checkmark$ & $\checkmark$                            \\
P7    & 0 - 3         & 300        & 1569          & $\checkmark$ & $\checkmark$ &                           &                           & $\checkmark$ &                           &                           &                                                      \\
P8    & 2 - 12        & 600        & 723           & $\checkmark$ & $\checkmark$ & $\checkmark$ & $\checkmark$ & $\checkmark$ &                           &                           &                         
\end{tblr}
\caption{A description of the ASAP Datasets: Scores, Average essay length in terms of words, and Traits. 
}
\label{tab:asap-dataset}
\end{table*}
\begin{table*}[h]
\centering
\setlength{\tabcolsep}{3.8pt}
\begin{tabular}{llccccccccc} 
\hline
Dataset                                      & Setup         & COH             & SYN             & VOC             & PHR             & GRM             & CVN             & HOL             &                 & Avg                      \\ 
\hline
\multicolumn{1}{c}{\multirow{6}{*}{ELLIPSE}} & Binary-1P     & 0.492           & 0.554           & 0.541           & 0.555           & 0.530           & 0.542           & 0.624           &                 & 0.548                    \\
\multicolumn{1}{c}{}                         & Binary-mP     & 0.515           & 0.575           & 0.580           & 0.594           & 0.548           & 0.559           & 0.647           &                 & 0.574                    \\
\multicolumn{1}{c}{}                         & Mutliclass-1P & 0.514           & 0.581           & 0.566           & 0.597           & \textbf{0.554}  & \textbf{0.577}  & 0.658           &                 & 0.578                    \\
\multicolumn{1}{c}{}                         & Multiclass-mP & \textbf{0.516}  & \textbf{0.583}  & \textbf{0.586}  & \textbf{0.603}  & 0.549           & 0.566           & \textbf{0.660}  &                 & \textbf{\textbf{0.581}}  \\ 
\cline{2-11}
\multicolumn{1}{c}{}                         & +PR           & 0.534$^\bullet$ & 0.589$^\bullet$ & 0.581           & 0.604$^\bullet$ & 0.557$^\bullet$ & 0.583$^\bullet$ & 0.663$^\bullet$ &                 & 0.587$^\bullet$          \\
\multicolumn{1}{c}{}                         & +PR +Features & 0.538$^\bullet$ & 0.590$^\bullet$ & 0.587$^\bullet$ & 0.610$^\bullet$ & 0.565$^\bullet$ & 0.585$^\bullet$ & 0.671$^\bullet$ &                 & 0.592$^\bullet$          \\ 
\hline\hline
                                             & Setup         & CNT             & ORG             & WC              & SF              & CNV             & PA              & LNG             & NAR             & Avg                      \\ 
\cline{2-11}
\multirow{6}{*}{ASAP}                        & Binary-1P     & 0.450           & 0.327           & 0.357           & 0.384           & 0.352           & 0.631           & 0.598           & 0.655           & 0.469                    \\
                                             & Binary-mP     & 0.429           & 0.355           & 0.392           & 0.427           & 0.357           & 0.608           & 0.587           & 0.641           & 0.474                    \\
                                             & Multiclass-1P & \textbf{0.625}  & \textbf{0.593}  & \textbf{0.531}  & \textbf{0.561}  & \textbf{0.588}  & \textbf{0.696}  & 0.664           & \textbf{0.718}  & \textbf{0.622}           \\
                                             & Multiclass-mP & 0.616           & 0.579           & 0.494           & 0.502           & 0.569           & 0.691           & \textbf{0.670}  & 0.703           & 0.603                    \\ 
\cline{2-11}
                                             & +PR           & 0.649$^\bullet$ & 0.604$^\bullet$ & 0.556$^\bullet$ & 0.602$^\bullet$ & 0.597$^\bullet$ & 0.704$^\bullet$ & 0.670$^\bullet$ & 0.727$^\bullet$ & 0.639$^\bullet$          \\
                                             & +PR +Features & 0.661$^\bullet$ & 0.626$^\bullet$ & 0.549           & 0.588           & 0.612$^\bullet$ & 0.713$^\bullet$ & 0.676$^\bullet$ & 0.737$^\bullet$ & 0.645$^\bullet$          \\ 
\hline\hline
                                             & Setup         & REL             & ORG             & VOC             & STY             & DEV             & MEC             & GRM             & HOL             & Avg                      \\ 
\cline{2-11}
\multirow{6}{*}{LAILA}                       & Binary-1P     & 0.306           & 0.594           & \textbf{0.668}  & 0.665           & 0.573           & 0.622           & 0.649           & \textbf{0.713}  & 0.599                    \\
                                             & Binary-mP     & 0.300           & \textbf{0.606}  & 0.666           & \textbf{0.669}  & \textbf{0.581}  & \textbf{0.626}  & \textbf{0.649}  & 0.709           & \textbf{0.601}           \\
                                             & Multiclass-1P & \textbf{0.308}  & 0.596           & 0.657           & 0.666           & 0.578           & 0.620           & 0.643           & 0.697           & 0.596                    \\
                                             & Multiclass-mP & 0.292           & 0.593           & 0.647           & 0.662           & 0.573           & 0.617           & 0.640           & 0.682           & 0.588                    \\ 
\cline{2-11}
                                             & +PR           & 0.303$^\bullet$ & 0.628$^\bullet$ & 0.680$^\bullet$ & 0.688$^\bullet$ & 0.606$^\bullet$ & 0.629$^\bullet$ & 0.654$^\bullet$ & 0.724$^\bullet$ & 0.614$^\bullet$          \\
                                             & +PR +Features & 0.275           & 0.662$^\bullet$ & 0.715$^\bullet$ & 0.706$^\bullet$ & 0.626$^\bullet$ & 0.638$^\bullet$ & 0.676$^\bullet$ & 0.747$^\bullet$ & 0.631$^\bullet$          \\
\hline
\end{tabular}
\caption{Performance of the meta-training setups in QWK on the \underline{dev sets}. Bold indicates the best-performing setup, and $^\bullet$ marks configurations that improved over the previous setting. PR indicates prompt and rubric.}
\label{tab:dev-results-full}
\end{table*}

\section{\sys{} Detailed Results} \label{sec:appendix-full}

This section presents the detailed results of the different meta-training setups at the trait level (Table \ref{tab:dev-results-full}) and analyzes how incorporating writing context (Section~\ref{sec:appendix-rq2}) and features (Section~\ref{sec:appendix-rq3}) affect the performance of \sys{}.

\subsection{Adding Writing Context}\label{sec:appendix-rq2}
We particularly investigate the effect of incorporating the prompt and rubric on the performance of \sys{}. Table \ref{tab:dev-results-full} shows that for ELLIPSE dataset, prompt-independent traits, e.g., cohesion, conventions, and grammar, benefit the most, with gains of 1-2 points, while prompt-dependent traits like vocabulary and phraseology show little or no improvement. This difference could be due to the extremely-short prompts of ELLIPSE (often just a couple of words), providing limited topical context for the prompt-dependent traits. In contrast, the rubric mainly benefits prompt-independent traits, which evaluate general writing quality.

On ASAP, traits with longer rubrics (e.g., sentence fluency and word choice) exhibit the largest gains, improving by 2-4 points, whereas traits with shorter rubrics (e.g., organization and narrativity) show only 1-point improvement.

Over LAILA, almost all traits improved. Prompt-independent traits, e.g., organization, style, and grammar, achieved the largest gains (2.2-2.5 points), while prompt-dependent traits like development and vocabulary improved less (1.4-1.8 points). Relevance showed little change, likely due to the narrow scoring range (0-2), which limits measurable improvement. The difference between ELLIPSE and LAILA may be attributed to the richer prompts of LAILA, allowing the model to better leverage both the rubric and the prompts.

\subsection{Adding Features}\label{sec:appendix-rq3}
A closer look at the rubrics explains why certain traits benefited more from incorporating the features. The added features primarily improve traits captured by surface-level and structural cues. For example, on ELLIPSE, traits aligned with text structure and correctness specifically, overall and grammar, improved the most, by nearly 1 point.
Moreover, on ASAP, organization, which emphasizes logical sequencing of ideas, and conventions, which covers proper grammar and punctuation, showed improvements of about 2 points.
In contrast, more substantial improvements were shown on LAILA, ranging from 1-4 points across most traits. The REL trait, however, was an exception, with performance decreasing by about 3 points. We note that relevance primarily depends on the semantic content of the essay, rather than syntactic or surface-level properties. As a result, adding feature-engineered representations, which focused more on structure, style, or readability, may have introduced noise rather than useful information for this trait.


\newpage
\section{\sys{} vs SOTA Detailed Results}\label{sec:full-results}
Since per-prompt comparison is a standard practice in AES literature, we report prompt-wise average performance of \sys{} on ELLIPSE, ASAP, and LAILA datasets in Tables~\ref{tab:ellipse-per-prompt}, \ref{tab:asap-per-prompt}, and \ref{tab:laila-per-prompt}, respectively. We note that for ASAP dataset, our results are not directly comparable to EPCTS \cite{EPCTS-2025}, as we do not predict the holistic trait in \sys{} for ASAP dataset (refer to Section \ref{sec:datasets}). Additionally, for ELLIPSE dataset, we report only MOOSE results (as its code is available), since \citet{eltanbouly2025trates} do not provide per-fold results for TRATES on ELLIPSE.

Overall, \sys{} outperforms the baselines on ELLIPSE and LAILA. On ASAP, it surpasses MOOSE but lags behind EPCTS (though the comparison is not direct). For ASAP dataset, the issue of non-unified score ranges is also evident in the per-prompt results: \sys{} exceeds MOOSE on P3, P4, P5, and P6, and even outperforms EPCTS on P3. However, it performs worse than MOOSE on P7 and P8.
Finally, for transparency, we detail the results of \sys{} on ELLIPSE, ASAP and LAILA datasets per-prompt and per-trait in Tables  \ref{tab:ellipse-test-full}, \ref{tab:asap-test-full} and \ref{tab:laila-test-full}, respectively. 

\begin{table}[ht]
\centering
\begin{tblr}{
rowsep=0pt,
  cells = {c},
  hline{1-2,13-14} = {-}{},
}
Fold & MOOSE  & MAPLE \\
1    & 0.248  & 0.571 \\
2    & 0.348  & 0.583 \\
3    & 0.036  & 0.647 \\
4    & 0.053  & 0.647 \\
5    & 0.343  & 0.601 \\
6    & 0.122  & 0.643 \\
7    & 0.336  & 0.594 \\
8    & 0.326  & 0.651 \\
9    & 0.391  & 0.685 \\
10   & 0.000  & 0.608 \\
11   & -0.004 & 0.665 \\
Avg  & 0.200  & 0.627 
\end{tblr}
\caption{\sys{} performance on ELLIPSE dataset per-fold averaged across all traits.}
\label{tab:ellipse-per-prompt}

\vspace{4em}

\centering
\begin{tblr}{
rowsep=0pt,
  cells = {c},
  hline{1-2,10-11} = {-}{},
}
Prompt & MOOSE & EPCTS* & MAPLE \\
P1     & 0.610 & 0.659  & 0.633 \\
P2     & 0.594 & 0.609  & 0.550 \\
P3     & 0.578 & 0.619~ & 0.698 \\
P4     & 0.619 & 0.686  & 0.681 \\
P5     & 0.477 & 0.671  & 0.648 \\
P6     & 0.509 & 0.629  & 0.591 \\
P7     & 0.382 & 0.555  & 0.315 \\
P8     & 0.483 & 0.630  & 0.360 \\
Avg    & 0.532 & 0.632  & 0.560 
\end{tblr}
\caption{\sys{} performance on ASAP dataset per-prompt averaged across all traits except HOL. EPCTS* is averaged over all traits, including the HOL trait, and is therefore not directly comparable.}
\label{tab:asap-per-prompt}

\vspace{4em}

\centering
\begin{tblr}{
rowsep=0pt,
  cells = {c},
  hline{1-2,10-11} = {-}{},
}
Prompt & MOOSE & XGB   & MAPLE \\
P1     & 0.426 & 0.400 & 0.537 \\
P2     & 0.652 & 0.656 & 0.601 \\
P3     & 0.661 & 0.710 & 0.679 \\
P4     & 0.625 & 0.564 & 0.612 \\
P5     & 0.442 & 0.549 & 0.500 \\
P6     & 0.661 & 0.653 & 0.691 \\
P7     & 0.658 & 0.561 & 0.664 \\
P8     & 0.646 & 0.653 & 0.668 \\
Avg    & 0.596 & 0.593 & 0.619 
\end{tblr}
\caption{\sys{} performance on LAILA dataset per-prompt averaged across all traits.}
\label{tab:laila-per-prompt}
\end{table}

\begin{table*}
\centering
\begin{tblr}{
rowsep=1pt,
  cells = {c},
  hline{1-2,13-14} = {-}{},
}
Fold & COH   & SYN   & VOC   & PHR   & GRA   & CON   & HOL   & Avg   \\
1    & 0.483 & 0.549 & 0.548 & 0.580 & 0.564 & 0.611 & 0.660 & 0.571 \\
2    & 0.567 & 0.521 & 0.611 & 0.615 & 0.565 & 0.559 & 0.644 & 0.583 \\
3    & 0.604 & 0.644 & 0.644 & 0.644 & 0.652 & 0.605 & 0.735 & 0.647 \\
4    & 0.628 & 0.668 & 0.625 & 0.656 & 0.604 & 0.615 & 0.735 & 0.647 \\
5    & 0.588 & 0.488 & 0.600 & 0.643 & 0.590 & 0.631 & 0.670 & 0.601 \\
6    & 0.590 & 0.667 & 0.595 & 0.649 & 0.641 & 0.639 & 0.720 & 0.643 \\
7    & 0.530 & 0.593 & 0.627 & 0.604 & 0.504 & 0.625 & 0.673 & 0.594 \\
8    & 0.600 & 0.657 & 0.626 & 0.647 & 0.627 & 0.676 & 0.724 & 0.651 \\
9    & 0.609 & 0.716 & 0.670 & 0.692 & 0.655 & 0.707 & 0.743 & 0.685 \\
10    & 0.533 & 0.606 & 0.575 & 0.623 & 0.597 & 0.657 & 0.664 & 0.608 \\
11   & 0.595 & 0.672 & 0.670 & 0.678 & 0.675 & 0.639 & 0.727 & 0.665 \\
Avg  & 0.575 & 0.616 & 0.617 & 0.639 & 0.607 & 0.633 & 0.700 & 0.627 
\end{tblr}
\caption{Breakdown of \sys{} prompt and trait-wise performance on ELLIPSE dataset.}
\label{tab:ellipse-test-full}
\end{table*}
\begin{table*}[h]
\centering
\begin{tblr}{
rowsep=1pt,
  cells = {c},
  hline{1-2,10-11} = {-}{},
}
Prompt & CNT   & ORG   & WC    & SF    & CNV   & PA    & LNG   & NAR   & Avg   \\
P1     & 0.447 & 0.648 & 0.682 & 0.705 & 0.684 & -     & -     & -     & 0.633 \\
P2     & 0.675 & 0.501 & 0.519 & 0.558 & 0.499 & -     & -     & -     & 0.550 \\
P3     & 0.693 & -     & -     & -     & -     & 0.696 & 0.679 & 0.725 & 0.698 \\
P4     & 0.677 & -     & -     & -     & -     & 0.700 & 0.619 & 0.727 & 0.681 \\
P5     & 0.682 & -     & -     & -     & -     & 0.643 & 0.617 & 0.648 & 0.648 \\
P6     & 0.548 & -     & -     & -     & -     & 0.561 & 0.617 & 0.638 & 0.591 \\
P7     & 0.373 & 0.309 & -     & -     & 0.262 & -     & -     & -     & 0.315 \\
P8     & 0.341 & 0.403 & 0.248 & 0.325 & 0.484 & -     & -     & -     & 0.360 \\
Avg    & 0.555 & 0.465 & 0.483 & 0.529 & 0.482 & 0.650 & 0.633 & 0.685 & 0.560 
\end{tblr}
\caption{Breakdown of \sys{} prompt and trait-wise performance on ASAP dataset.}
\label{tab:asap-test-full}
\end{table*}
\begin{table*}[h]
\centering
\begin{tblr}{
rowsep=1pt,
  cells = {c},
  hline{1-2,10-11} = {-}{},
}
Prompt & REL   & ORG   & VOC   & STY   & DEV   & MEC   & GRA   & HOL   & Avg   \\
P1     & 0.270 & 0.578 & 0.608 & 0.632 & 0.553 & 0.515 & 0.527 & 0.613 & 0.537 \\
P2     & 0.392 & 0.612 & 0.650 & 0.687 & 0.547 & 0.614 & 0.622 & 0.682 & 0.601 \\
P3     & 0.216 & 0.690 & 0.716 & 0.792 & 0.665 & 0.752 & 0.791 & 0.814 & 0.679 \\
P4     & 0.365 & 0.601 & 0.579 & 0.662 & 0.561 & 0.660 & 0.673 & 0.791 & 0.612 \\
P5     & 0.217 & 0.588 & 0.592 & 0.569 & 0.528 & 0.477 & 0.516 & 0.513 & 0.500 \\
P6     & 0.182 & 0.769 & 0.767 & 0.800 & 0.763 & 0.653 & 0.742 & 0.853 & 0.691 \\
P7     & 0.277 & 0.668 & 0.791 & 0.726 & 0.661 & 0.700 & 0.736 & 0.751 & 0.664 \\
P8     & 0.401 & 0.672 & 0.784 & 0.750 & 0.574 & 0.697 & 0.706 & 0.763 & 0.668 \\
Avg    & 0.290 & 0.647 & 0.686 & 0.702 & 0.607 & 0.634 & 0.664 & 0.723 & 0.619 
\end{tblr}
\caption{Breakdown of \sys{} prompt and trait-wise performance on LAILA dataset.}
\label{tab:laila-test-full}
\end{table*}



\end{document}